\documentclass[runningheads]{llncs}
\usepackage{amsmath,amssymb} 
\usepackage{tikz}
\usepackage{graphicx}
\usepackage{comment}
\usepackage{color}

\usepackage{hyperref}
\usepackage{epsfig}
\usepackage{algorithm2e}
\usepackage[normalem]{ulem}
\usepackage{subfigure}
\usepackage{mathtools}
\usepackage{wrapfig}

\def\onedot{. }
\def\eg{\emph{e.g.},} 
\def\ie{\emph{i.e.},}

\def\etal{\emph{et al}\onedot}

\def\eg{\emph{e.g.},} 

\newcommand{\Tref}[1]{Table~\ref{#1}}
\newcommand{\Eref}[1]{Eq.~(\ref{#1})}
\newcommand{\Fref}[1]{Fig.~\ref{#1}}
\newcommand{\Aref}[1]{Alg.~\ref{#1}}

\newcommand{\Cref}[1]{Chap.~\ref{#1}}
\newcommand{\Sref}[1]{Sec.~\ref{#1}}

\begin{document}
\pagestyle{headings}
\mainmatter
\def\ECCVSubNumber{3850}  

\title{Detecting Human-Object Interactions with Action Co-occurrence Priors} 

\titlerunning{Action Co-occurrence Priors}
%
\author{%
  Dong-Jin Kim\inst{1}\quad%
  Xiao Sun\inst{2}\quad%
  Jinsoo Choi\inst{1}\quad%
  Stephen Lin\inst{2}\quad%
  In So Kweon\inst{1}
  %
  }
\authorrunning{D. Kim et al.}
\institute{KAIST, South Korea. 
\email{\{djnjusa,jinsc37,iskweon77\}@kaist.ac.kr}\and
Microsoft Research, China.
\email{\{xias,stevelin\}@microsoft.com}
\url{{https://github.com/Dong-JinKim/ActionCooccurrencePriors/}}
}

%

%

\maketitle
\begin{abstract}
A common problem in human-object interaction (HOI) detection task is that numerous HOI classes have only a small number of labeled examples, resulting in training sets with a long-tailed distribution. 
The lack of positive labels can lead to low classification accuracy for these classes. 
Towards addressing this issue, we observe that there exist natural correlations and anti-correlations among human-object interactions. 
In this paper, we model the correlations as {\em action co-occurrence matrices} and present techniques to learn these priors and leverage them for more effective training, especially on rare classes.
The utility of our approach is demonstrated experimentally, where the performance of our approach exceeds the state-of-the-art methods on both of the two leading HOI detection benchmark datasets, HICO-Det and V-COCO.
\end{abstract}

\section{Introduction}
Human-object interaction (HOI) detection aims to localize humans and objects in an image and infer the relationships between them. 
An HOI is typically represented as a human-action-object triplet with the corresponding bounding boxes and classes. 
Detecting these interactions is a fundamental challenge in visual recognition that requires both an understanding of object information and high-level knowledge of interactions.


\begin{figure}[t]
	\centering
	\vspace{0mm}
	\includegraphics[width=.6\linewidth,keepaspectratio]{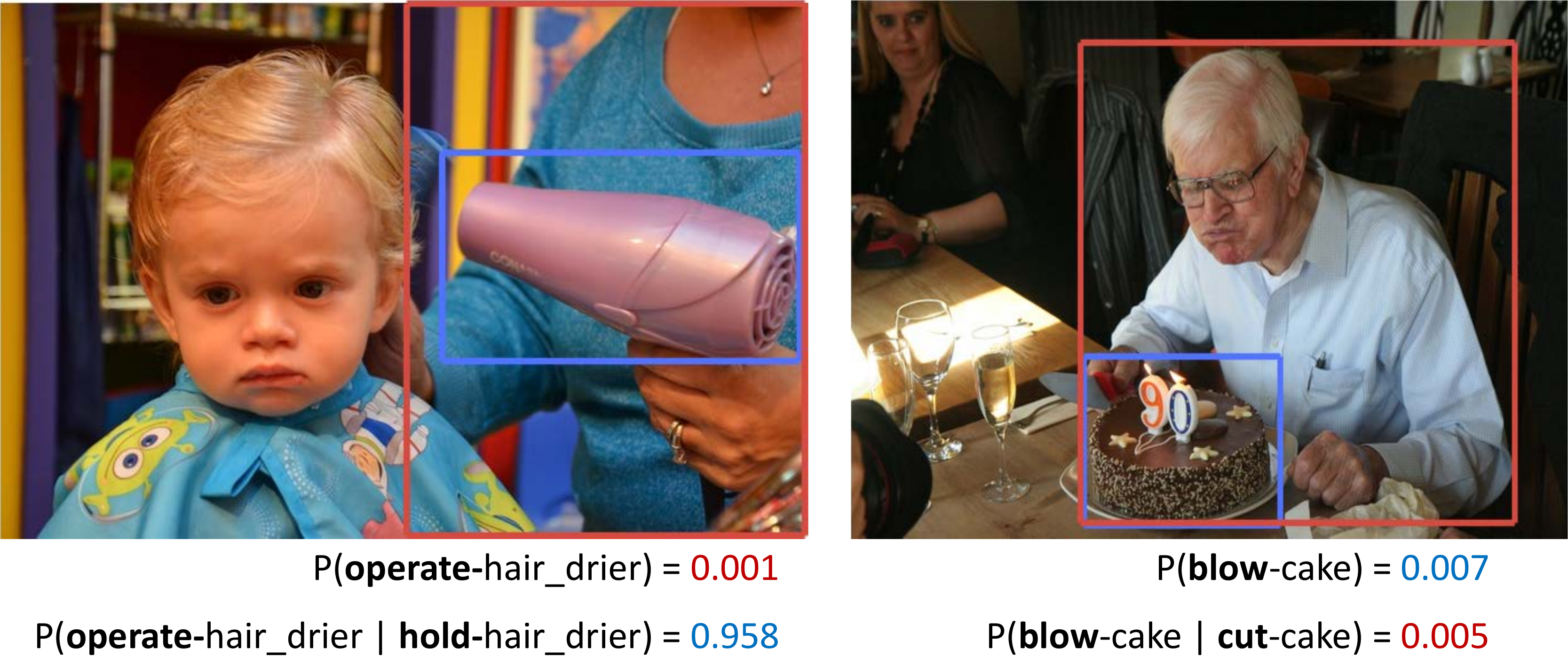}
	\vspace{-4mm}
	\caption{Examples of action co-occurrence. 
	{The marginal/conditional probability values are computed from the distribution of the training label. Intuitively,} detection of rarely labeled HOIs (operate-hair dryer) can be facilitated by detection of commonly co-occurring HOIs (hold-hair dryer). Also, non-detection of rare HOIs (blow-cake) can be aided by detection of incompatible HOIs (cut-cake). 
	We leverage this intuition as a prior to learn an HOI detector effective on long-tailed datasets.
	}
	\label{fig:teaser}
	\vspace{-6mm}
	
\end{figure}

A major issue that exists in HOI detection is that its datasets suffer from long-tailed distributions, in which many HOI triplets have few labeled instances.
Similar to datasets for general visual relationship detection (VRD)~\cite{lu2016visual}, a reason for this is missing labels, where the annotation covers only a subset of the interactions present in an image.
For the widely-used HICO-Det dataset~\cite{chao2018learning}, 462 out of the 600 HOI classes have fewer than 10 training samples. 
For such classes, the lack of positive labels can lead to inadequate training and 
low classification performance.
{How to alleviate the performance degradation on rare classes is thus a key issue in HOI detection.}


To address the problem of long-tailed distributions, we propose to take advantage of natural \emph{co-occurrences} in human actions. For example, the HOI of `operate-hair dryer' is rarely labeled and consequently hard to detect in the left image of \Fref{fig:teaser}. 
However, `operate-hair dryer' often occurs when the more commonly labeled HOI of `hold-hair dryer' is present. As a result, detection of `operate-hair dryer' can be facilitated by detection of `hold-hair dryer' in an image. On the other hand, the detection of an HOI may preclude other incompatible HOIs, such as for `cut-cake' and `blow-cake' in the right image of \Fref{fig:teaser}.

In this paper, we introduce {the new concept of utilizing co-occurring actions as prior knowledge, termed as} action co-occurrence priors (ACPs), to train an HOI detector. 
{In contrast to the language-based prior knowledge which requires external data sources~\cite{kim2019image,lu2016visual,yu2017visual}, the co-occurrence priors can be easily obtained from the statistics of the target dataset.}
{We also propose} two novel ways to exploit them. 
First, we {design} a {neural network with hierarchical structure} where the classification is initially performed with respect to {\em action groups}.
{Each action group is defined by one anchor action, where the anchor actions are mutually exclusive according to the co-occurrence prior.}
{Then, our model predicts the fine-grained HOI class within the action group.}
{Second, we present a technique that employs knowledge distillation~\cite{hinton2015distilling} to expand HOI labels 
so they can have more positive labels for potentially co-occurring actions.} 
{During training, the predictions are regularized 
by the refined objectives
to improve robustness, especially for classes in the data distribution tail.}
To the best of our knowledge, we are the first to leverage the label co-occurrence in HOI detection to alleviate long-tailed distribution.


{The main contributions of this work can be summarized as: 
(1) The novel concept of explicitly leveraging correlations among HOI labels to address the problem of long-tailed distributions in HOI detection; 
(2) Two orthogonal ways to leverage action co-occurrence priors, namely through a proposed hierarchical architecture and HOI label expansion via knowledge distillation.
The resulting model is shown to be consistently advantageous in relation to state-of-the-art techniques on both the HICO-Det~\cite{chao2018learning} and V-COCO~\cite{gupta2015visual} benchmark datasets.} 




\section{Related Work}


\noindent\textbf{Human-Object Interaction} 
Human-Object Interaction was originally studied in the context of recognizing the function or `affordance' of objects~\cite{delaitre2012scene,gibson2014ecological,grabner2011makes,gupta2007objects,stark1991achieving}. Early works focus on learning more discriminative features combined with variants of SVM classifiers~\cite{delaitre2010recognizing,delaitre2011learning,yao2010grouplet}, and leverage the relationship with human poses for better representation~\cite{delaitre2010recognizing,delaitre2011learning,yao2011human} or mutual context modeling~\cite{yao2010modeling}.


Recently, a completely data-driven approach based on convolutional neural networks (CNNs) has brought dramatic progress to HOI. Many of the pioneering works created large scale image datasets~\cite{chao2018learning,chao2015hico,gupta2015visual,zhuang2018hcvrd} 
to set new benchmarks in this field. 
Henceforth, significant progress have been seen in using CNNs for this problem~\cite{bansal2020detecting,chao2018learning,gao2018ican,gkioxari2018detecting,gupta2019no,kato2018compositional,li2020detailed,li2019transferable,liao2019ppdm,qi2018learning,shen2018scaling,ulutan2020vsgnet,wan2019pose,wang2020learning,xu2019interact,xu2019learning}. 
Most of these works follow a two-step scheme of CNN feature extraction and multi-information fusion, where the multiple information may include 
human and object appearance~\cite{chao2018learning,gao2018ican,gkioxari2018detecting,qi2018learning}; box relation (either box configuration or spatial map)~\cite{bansal2020detecting,chao2018learning,gkioxari2018detecting,gupta2019no,xu2019learning}; object category~\cite{bansal2020detecting,gupta2019no,peyre2019detecting}; human pose~\cite{gupta2019no,li2020detailed,li2019transferable};
and particularly, linguistic prior knowledge~\cite{kato2018compositional,peyre2019detecting}. 
More recent works tend to combine these various cues~\cite{gupta2019no,li2019transferable,wan2019pose,xu2019interact}.
These works differ from one another mainly in their techniques for exploiting external knowledge priors. 
Kato~\etal~\cite{kato2018compositional} incorporate information from WordNet~\cite{miller1995wordnet} using a Graph Convolutional Network (GCN)~\cite{kipf2016semi} and learn to compose new HOIs. 
Xu~\etal~\cite{xu2019learning} also use a GCN to model the general dependencies among actions and object categories by leveraging a VRD dataset~\cite{lu2016visual}. 
Li~\etal~\cite{li2019transferable} utilize interactiveness knowledge learned across multiple HOI datasets. 
Peyre~\etal~\cite{peyre2019detecting} transfer knowledge from triplets seen at training to new unseen triplets at test time by analogy reasoning.




Different from the previous works, our approach is to reformulate the target action label space and corresponding loss function in a manner that leverages \emph{co-occurrence} relationships among action classes for HOI detection. In principle, the proposed technique is complementary to all of the previous works and can be combined with any of them. For our experiments, we implemented our approach on a baseline presented in~\cite{gupta2019no}, with details given in \Sref{sec.architecture}.






\noindent\textbf{Visual Relationship Detection} The closest problem to HOI detection is Visual Relationship Detection (VRD)~\cite{dai2017detecting,li2017vip,plummer2016phrase,yang2018shuffle,yin2018zoom,yu2017visual,zhan2019exploring,zhang2017visual,zhang2017relationship,zhuang2017towards}, which deals with general visual relationships between two arbitrary objects. 
In the VRD datasets~\cite{krishna2017visual,lu2016visual}, the types of visual relationships that are modeled include verb (action), preposition, spatial and comparative phrase. 
The two tasks share common challenges such as long-tail distributions or even zero-shot problems~\cite{lu2016visual}. 
Our work focuses on HOI detection, as co-occurrences of human-object interactions are often strong, but the proposed technique could be extended to model the general co-occurrences that exist in visual relationships.





\noindent\textbf{Label Hierarchy in Multi-label Learning} The hierarchical structure of label categories has long been exploited for multi-label learning in various vision tasks, \eg~image/object classification~\cite{deng2014large,yan2015hd}, detection~\cite{hwang2011sharing,marszalek2007semantic}, and human pose estimation~\cite{sun2019explicit,sun2015cascaded}. In contrast, label hierarchy has rarely been considered in HOI detection. Inspired by previous work~\cite{deng2014large} that uses Hierarchy and Exclusion (HEX) graphs to encode flexible relations between object labels, we present the first method to take advantage of an action label hierarchy for HOI recognition.
{While label hierarchies have commonly been used, our method is different in that it is defined by co-occurrences (rather than semantics or a taxonomy~\cite{deng2014large}). This co-occurrence based hierarchy can be determined statistically, without direct human supervision.}

\section{Proposed Method}
\vspace{-2mm}
{Our method for utilizing the co-occurrence information of HOI labels consists of three key components:} 
(1) establishing action co-occurrence priors (\Sref{sec.prior}), (2) hierarchical learning including anchor action selection (\Sref{sec.anchor}) and devising the hierarchical architecture (\Sref{sec.architecture}), and 
(3) 
{ACP projection for knowledge distillation} (\Sref{sec.knowledge_distillation}).

\subsection{Action Co-occurrence Priors}
\label{sec.prior}

\begin{figure*}[t]
    \centering
    \vspace{0mm}
    \includegraphics[width=0.32\linewidth,keepaspectratio]{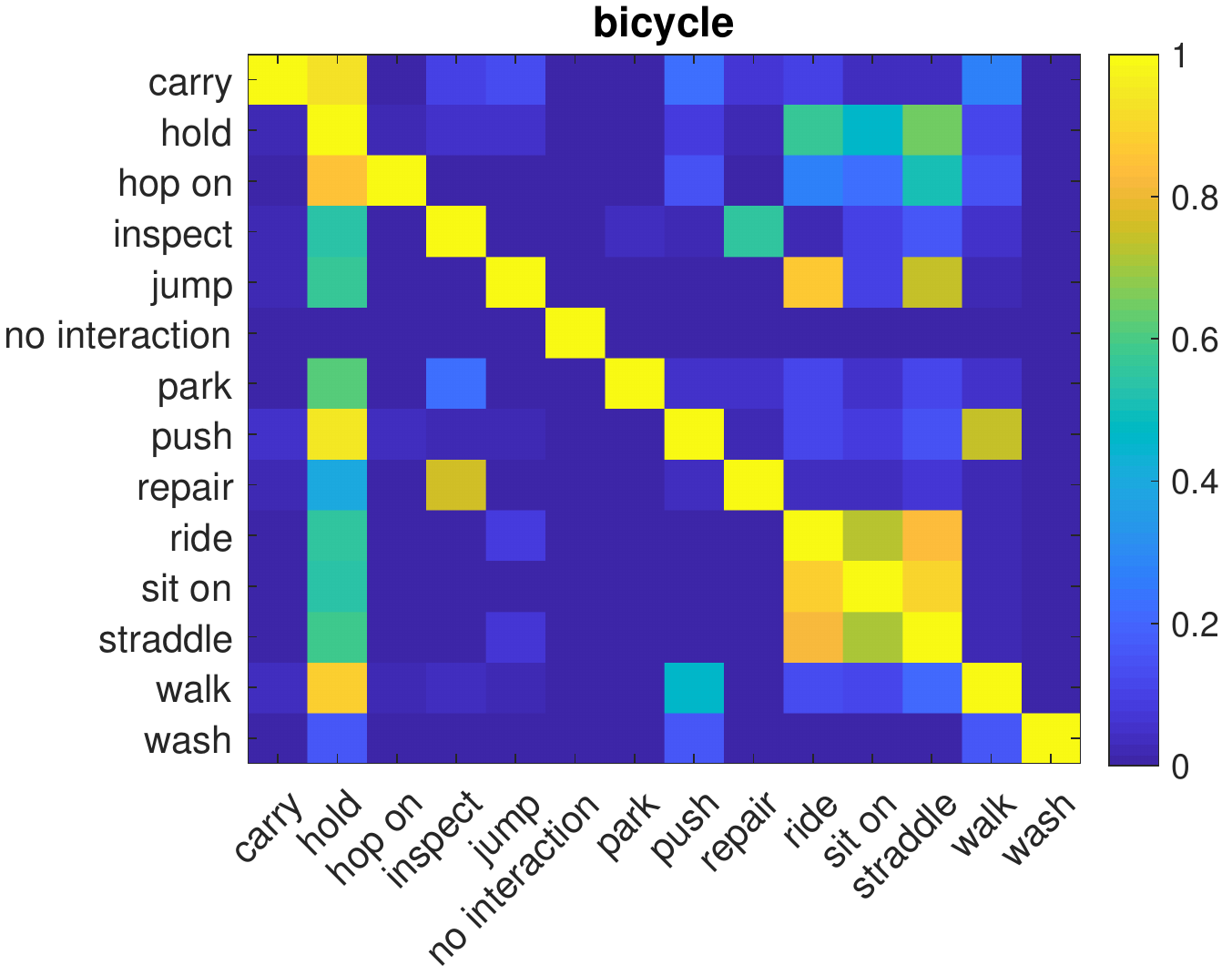}
    \includegraphics[width=0.32\linewidth,keepaspectratio]{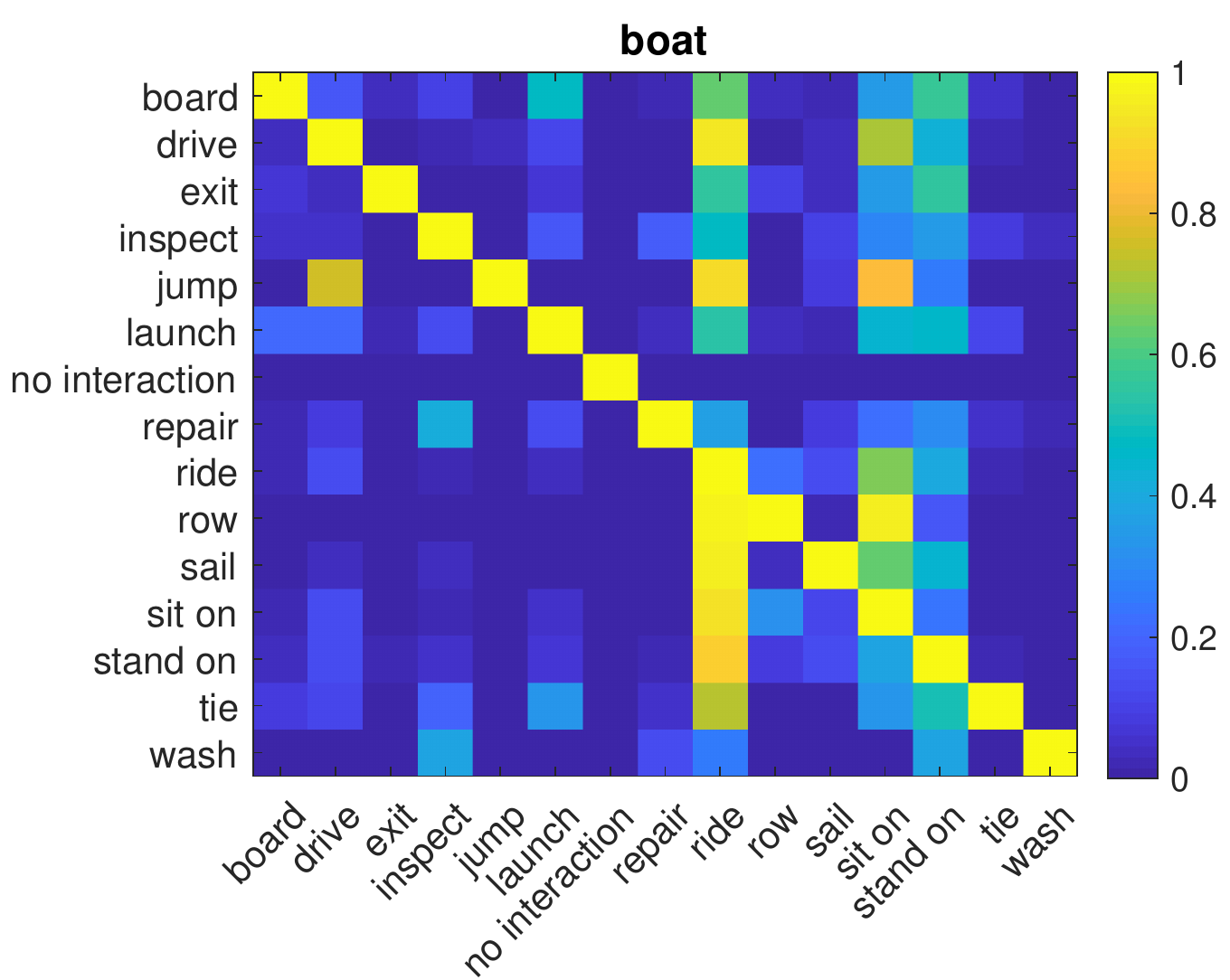}
    \includegraphics[width=0.32\linewidth,keepaspectratio]{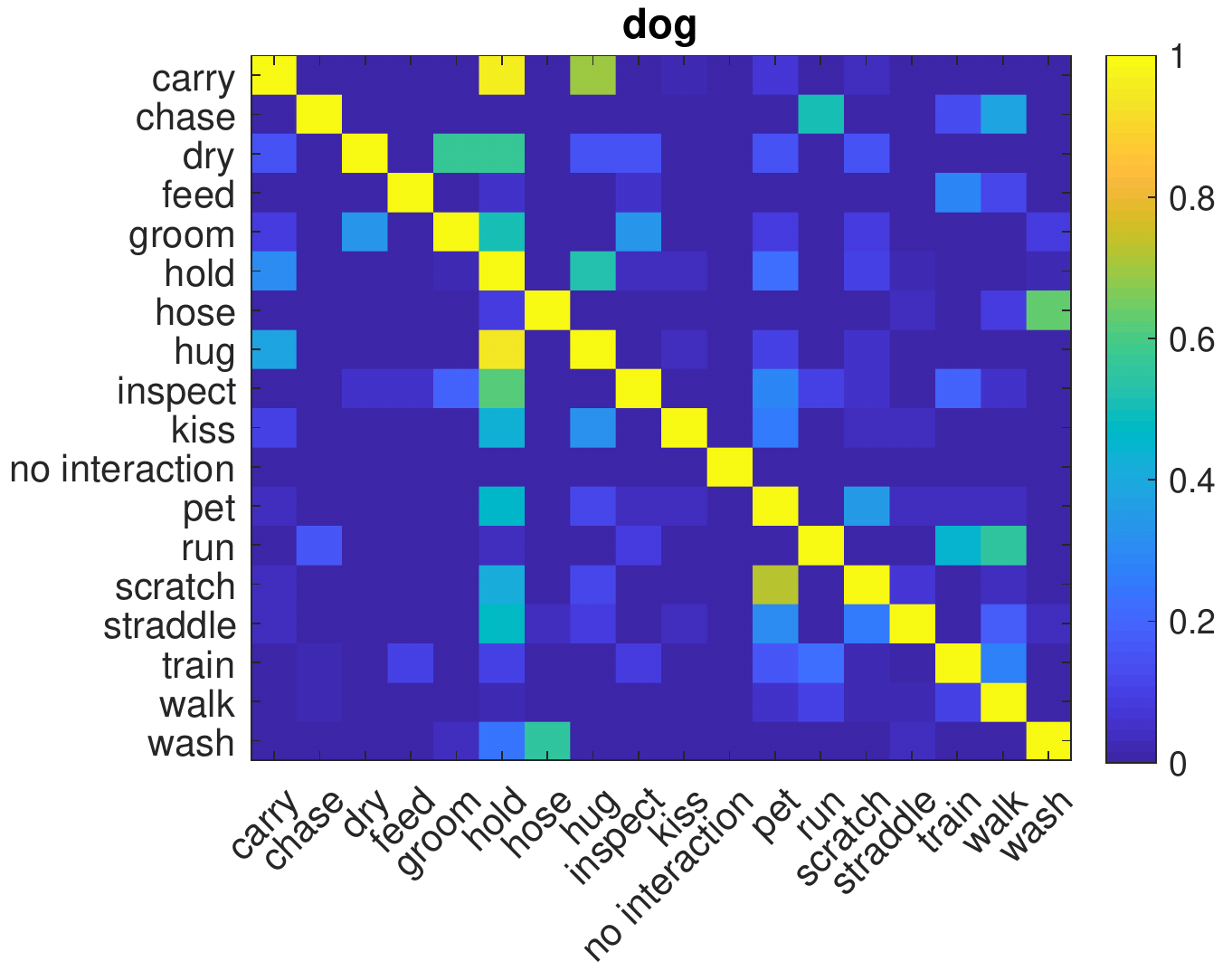}
    \vspace{-4mm}
    \caption{Examples of co-occurrence matrices constructed for several objects (bicycle, boat, dog). 
    Along the Y-axis is the given action, and the X-axis enumerates conditional actions. 
    Each element represents the conditional probability that an action occurs when another action is happening.
    }
    \vspace{-6mm}
    \label{fig:co-occurrence}
\end{figure*}

{Here, we formalize the action co-occurrence priors.} 
The priors for the actions are modeled by a co-occurrence matrix $C \in \mathbb{R}^{N\times N}$ where an entry $c_{ij}$ in $C$ represents the conditional probability that action $j$ occurs when action $i$ is happening: 
\begin{equation}
    c_{ij} = p(j|i), i,j \in [0, N),
\label{eq:acp}
\end{equation}
where $N$ 
denotes the total number of actions classes and $i,j$ are indices of two actions. 
$C$ is constructed from the target HOI detection dataset by counting the {image-level} statistics of its training labels.
Examples of co-occurrence matrices constructed for single object are visualized in \Fref{fig:co-occurrence}.




Meanwhile, we also consider the complementary event of action $i$ (\ie~where the $i$-th action does not occur) and denote it as $i^{'}$, such that $p(i^{'}) + p(i) = 1$.
The complementary action co-occurrence matrix $C^{'} \in \mathbb{R}^{N\times N}$ can thus be defined by entries $c_{ij}^{'}$ in $C^{'}$ that represent the conditional probability that an action $j$ occurs when another action $i$ does not occur: 
\begin{equation}
    c^{'}_{ij} = p(j|i^{'}), i,j \in [0, N).
\end{equation}



It can be seen from \Fref{fig:co-occurrence}, that different types of relationships can exist between actions. 
The types can be divided into three types.
The first type is the \textbf{prerequisite} relationship, where the given action is highly likely to co-occur with the conditional action. 
For example, the HOI `sit on-bicycle' is a prerequisite of the HOI `ride-bicycle'. 
In this case, 
$p(\text{sit on-bicycle}|\text{ride-bicycle})$ is close to $1$.
Next, \textbf{exclusion}, where the given action is highly unlikely to co-occur with the conditional action. An example is that the HOI `wash-bicycle' and HOI `ride-bicycle' are unlikely to happen together. 
As a result, $p(\text{wash-bicycle}|\text{ride bicycle})$ is close to $0$.
Finally, \textbf{overlapping}, where the given action and conditional action may possibly co-occur, for example HOI `hold-bicycle' and HOI `inspect-bicycle', such that $p(\text{hold-bicycle}|\text{inspect-bicycle})$ is in between $0$ and $1$.

The strong relationships that may exist between action labels can provide strong priors on the presence or absence of an HOI in an image. 
{In contrast to previous works where models may implicitly learn label co-occurrence via relational architectures~\cite{baradel2018object,zhang2019co}, we explicitly exploit these relationships between action labels as priors, to effectively train our model especially for rare HOIs.}

\subsection{Anchor Action Selection via Non-Exclusive Action Suppression}
\label{sec.anchor}
{{From a co-occurrence matrix for an object, it can be seen that some actions are close in semantics or commonly co-occur while others are not. }}
Intuitively, closely related actions (\eg~`sit on-bicycle' and `straddle-bicycle') tend to be harder to distinguish from each other.
{If the positive labels for these actions are rare, then they become even more difficult to distinguish.}
Such cases require fine-grained recognition~\cite{fu2017look} and demand more dedicated classifiers.
This motivates us to learn HOI classes in a coarse-to-fine manner. Specifically, we first identify a set of mutually exclusive action {classes}, called \emph{anchor actions}, which tend to be distinguishable from one another. The anchor actions will be used to partition the entire action {label} space into fine-grained sub-spaces. The other action {classes} will be attributed to one or more sub-spaces and recognized in the context of a specific anchor action.
{In summary, unlike previous HOI detection works which predict action probabilities independently of one another, we divide the whole action label set into two sets, one for anchor actions and one for regular actions, which are modeled in different ways as explained in detail in \Sref{sec.architecture}}.
In selecting anchor actions, we seek a set of action {classes} that are exclusive of one another. Toward this end, we define the exclusiveness of an action {class} as counting the number of actions that will never occur if action $i$ is happening ($e_i = \sum_j ({1} \text{ if } (c_{ij} =0), \text{ else } {0})$).
$e_i$ will have a high value if few other actions can occur when $i$ does.
Based on exclusiveness values, the anchor action {label} set $\mathcal{D}$ is generated through \emph{\textbf{non-exclusive suppression (NES)}} as described in \Aref{alg:correlation}. It iteratively finds the most exclusive action {class} as an anchor action and removes remaining action {classes} that are not exclusive to the selected anchor actions. 
The anchors in the list are action {classes} that never occur together in the training labels. For example, if an action such as `toast' is on the anchor list, then actions like `stand' and `sit' cannot be on the list because they may co-occur with `toast', while actions such as `hunt' or `hop on' can potentially be on the list.
{While there may exist other ways the anchor action selection could be done, we empirically found this approach to be simple, effective (detection accuracy), and efficient (less than 0.01 second). }

The anchor action {label} set acts as a finite partition of the action {label} space (a set of pairwise disjoint events whose union is the entire action {label} space). To form a complete action {label} space, we add an \emph{\textbf{`other'}} anchor action, denoted as $\mathcal{O}$, for when an action {class} does not belong to $\mathcal{D}$. 
Finally, we get $|\mathcal{D}| + 1$ anchor actions including $\mathcal{D}$ and the `other' action {class} $\mathcal{O}$. 

\begin{wrapfigure}{R}{0.55\textwidth}
\vspace{-8mm}
\resizebox{1\linewidth}{!}{
\begin{minipage}{0.7\textwidth}

\begin{algorithm}[H]
\KwIn{
$\mathcal{E}=\{e_{i}, i \in [0, N)\}, \mathcal{C}=\{c_{ij}, i,j \in [0, N)\}$\;
}
\KwOut{
$\mathcal{D}$
} 
\Begin
{
  $\mathcal{D} \gets \{\}$  \;
  \While{$\mathcal{E}$ is not empty}
  {
    \# Find the most exclusive action class\;
    $m \gets argmax \mathcal{E}$\;
   $\mathcal{D} \gets \mathcal{D} \cup \{m\}$  \;
    \For{$e_k \in \mathcal{E}$}{
            \# Remove the action classes correlated (not exclusive) to $m$\;
            \If{$c_{mk} > 0$}
            {
                $\mathcal{E} \gets \mathcal{E} - e_k; \mathcal{C} \gets C-\{c_{ij}, i \text{ or } j = k\}$
            }
    }
  }
}
\label{alg:correlation}
\caption{
{Non-Exclusive Suppression (NES) algorithm for mutually exclusive anchor action selection.
}
}
\end{algorithm}
\end{minipage}
}
\vspace{-8mm}
\end{wrapfigure}

There are several benefits to having this anchor action {label} set.
First, only one anchor action can happen at one time between a given human and object. Thus, we can use the relative (one hot) probability representation with \emph{softmax} activation, whose features were shown to compare well against distance metric learning-based features~\cite{horiguchi2019significance}.
Second, anchor actions tend to be easier to distinguish from one another since they generally have prominent differences in an image.
Third, it decomposes the action {label} space into several sub-spaces, which facilitates a coarse-to-fine solution. Each sub-task will have a much smaller solution space, which can improve learning.
Finally, each sub-task will use a standalone sub-network which focuses on image features specific to the sub-task, which is an effective strategy for fine-grained recognition~\cite{fu2017look}.

After selecting anchor actions, the entire action {label} set $\mathcal{A}$ is divided into the anchor action {label} set $\mathcal{D}$ and the remaining set of `regular' action {classes} $\mathcal{R}$, so that $\mathcal{A} = \{\mathcal{D}, \mathcal{R}\}$.
Each of the regular action {classes} is then associated with one or more anchor actions to form $|\mathcal{D}|+1$ action groups $\mathbf{G} = \{\mathcal{G}_i; i \in \mathcal{D}\cup\mathcal{O}\}$, one for each anchor action. 
A regular action {class} $j \in \mathcal{R}$ will be assigned to the group of anchor action $i$ ($\mathcal{G}_i$) if action $j$ is able to co-occur with the anchor action $i$,
\begin{equation}
    j \in \mathcal{G}_i,  \text{if } c_{ij} > 0 \;\; (i \in \mathcal{D}\cup\mathcal{O}, j \in \mathcal{R}).
\label{eq:action_groups2}
\end{equation}
Note that the anchor actions themselves are not included in the action groups and a regular action $j$ can be assigned to multiple action groups since it may co-occur with multiple anchor actions.



\subsection{Hierarchical Architecture}
\label{sec.architecture}

We implemented our hierarchical approach upon the `No-Frills' (NFs) baseline presented in~\cite{gupta2019no} on account of its simplicity, effectiveness, and code availability~\cite{nofrillsgit}. 
Here, we give a brief review of the NFs architecture. 

\begin{figure}[t]
\centering
    \vspace{0mm}
    \includegraphics[width=1\linewidth,keepaspectratio]{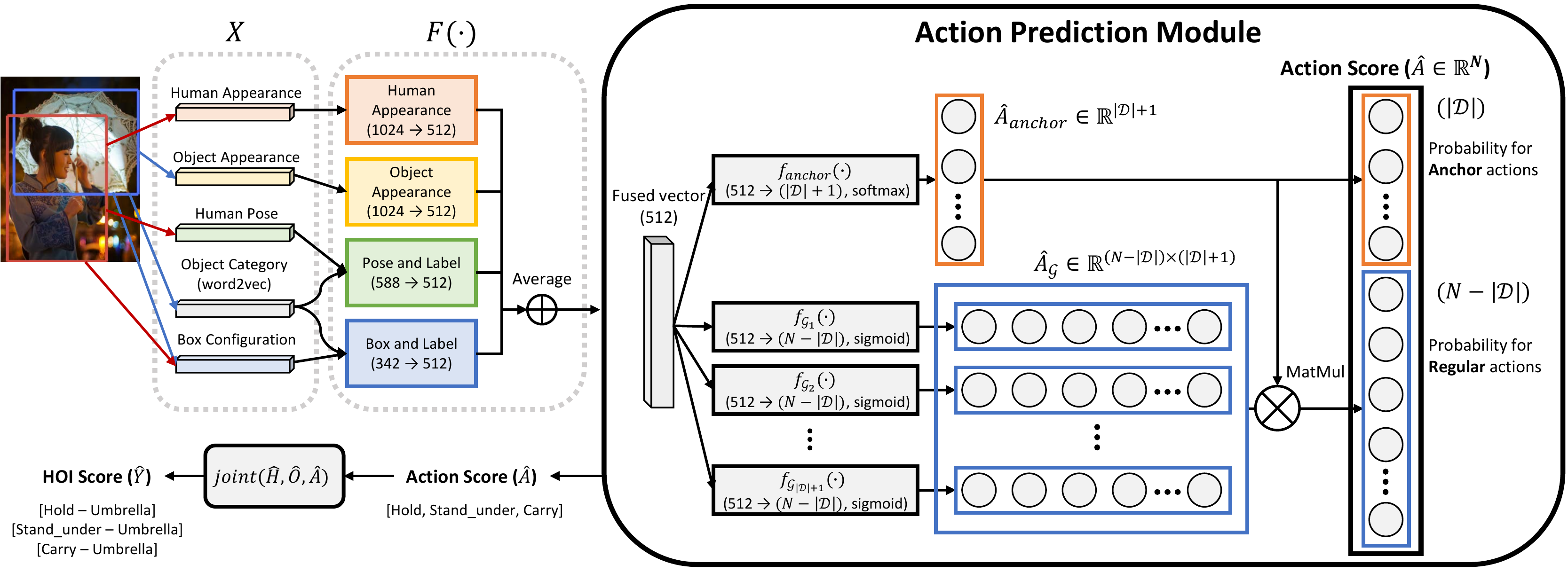}\\
    \vspace{-2mm}
    \caption{Illustration of our overall network architecture. 
	Our work differs from the baseline \cite{gupta2019no} by the addition of a hierarchical \emph{action prediction module}.
	For our hierarchical architecture, anchor action probability is directly generated by a \emph{softmax} sub-network. Regular action probability is generated by a matrix multiplication of the anchor probability and the output from a few \emph{sigmoid} based conditional sub-networks.}
    \label{fig:model_merged}
    \vspace{-6mm}
\end{figure}

\noindent\textbf{Baseline Network}
NFs follows the common scheme of CNN feature extraction followed by multi-information fusion. It uses the off-the-shelf Faster R-CNN~\cite{ren2015faster} object detector with ResNet152~\cite{he2016deep} backbone network to detect human and object bounding boxes. 
As illustrated in \Fref{fig:model_merged}, the multiple information used in~\cite{gupta2019no} (denoted as $X$) are fed through four separate network streams to generate fixed dimension features. Then, all the features are added together and sent through a \emph{sigmoid} activation to get the action probability prediction $\hat{A}$:
\begin{equation}
    \hat{A}= sigmoid(F(X)) \in \mathbb{R}^N,
\label{eq:base_action_pred}
\end{equation}
where $\hat{A}(a) = p(a|X)$ represents the probability prediction for action class $a$.


To eliminate training-inference mismatch, NFs directly optimizes the HOI class probabilities instead of separating the detection and interaction losses as done in~\cite{gao2018ican,gkioxari2018detecting}. The final HOI prediction is a joint probability distribution over $M$ number of HOI classes {computed from the probabilities $\hat{H}$, $\hat{O}$, and $\hat{A}$ for human, object, and action, respectively:}
\begin{equation}
    \hat{Y} = joint(\hat{H}, \hat{O}, \hat{A}) \in \mathbb{R}^{M}.
\end{equation}
Specifically, for a HOI class $(h,o,a)$,
\begin{equation}
    \hat{Y} (h,o,a)  = \hat{H}(h) * \hat{O}(o) * \hat{A}(a)
     = p(h|I) * p(o|I) * p(a|X),\\
    \label{eq:base_hoi_pred}
\end{equation}
where $\hat{H}(h)=p(h|I)$ and $\hat{O}(o)=p(o|I)$ are the probability of a candidate box pair being a human $h$ and object $o$, provided by the object detector~\cite{ren2015faster}. Finally, the binary cross-entropy loss $\mathcal{L}(\hat{Y}, Y^{gt})$ is directly computed from the HOI prediction. 
This ‘No-Frills’ baseline network is referred to as \emph{\textbf{Baseline}}.



\noindent\textbf{Modified Baseline Network} 
{For a stronger baseline comparison, }
we make two {simple} but very effective modifications on the baseline network.
(1) Replace the one-hot representation 
with the Glove word2vec~\cite{pennington2014glove} representation for the object category.
(2) Instead of directly adding up the multiple information, we average them and forward this through another \emph{action prediction module}
    to obtain the final action probability prediction. For a naive approach (the \emph{\textbf{Modified  Baseline}}), we simply use a sub-network $f_{sub}$  of a few FC layers as the \emph{action prediction module}.
    Then \Eref{eq:base_action_pred} is modified to
    \begin{equation}
     \hat{A}=     sigmoid(f_{sub}(F(X))).
     \label{eq:modified}
    \end{equation}



Our hierarchical architecture further modifies the \emph{action prediction module} by explicitly exploiting ACP information which is described in the next paragraph.

\noindent\textbf{Proposed Hierarchical Architecture}
Now we introduce the \emph{action prediction module} for our hierarchical architecture (illustrated in \Fref{fig:model_merged}) that better exploits the inherent co-occurrence among actions. While the baseline network predicts all the action probabilities directly from $F(\cdot)$ with a single feed-forward sub-network $f_{sub}$, we instead use $|\mathcal{D}|+2$ sub-networks where one ($f_{anchor}(\cdot)$) is first applied to predict the anchor action set and then one of the $|\mathcal{D}|+1$ other sub-networks ($f_{\mathcal{G}_i}(\cdot)$) which corresponds to the predicted anchor action is used to estimate the specific action within the action group. Because of the mutually exclusive property of anchor actions, we use the \emph{softmax} activation for anchor action predictions, while employing the \emph{sigmoid} activation for regular action predictions conditional to the action groups:

\begin{equation}
    \hat{A}_{anchor} = softmax(f_{anchor}(F(X))) \in \mathbb{R}^{|\mathcal{D}|+1}
\label{eq:hie_anchor_pred}
\end{equation}
\begin{equation}
    \hat{A}_{\mathcal{G}_i} = sigmoid(f_{\mathcal{G}_i}(F(X))) \in \mathbb{R}^{N-|\mathcal{D}|}, \text{where } i \in \mathcal{D}\cup\mathcal{O},
\label{eq:hie_group_pred}
\end{equation}
where $\hat{A}_{anchor}(i)$ is directly used as the final probability predictions for the anchor actions ($p(i|X) = \hat{A}_{anchor}(i), i \in \mathcal{D}$).
We let $\hat{A}_{\mathcal{G}_i}(j)$ represent the learned conditional probability that action $j$ occurs when action $i$ is happening
($p(j|i,X) = \hat{A}_{\mathcal{G}_i}(j)$).
Since the anchor action set is a finite partition of the entire action {label} space, the probability of a regular action $j$ can be predicted according to the law of total probability:
\begin{equation}
    \footnotesize
    \hat{A}_{regular}(j)  = p(j|X)
    =\sum_{ i \in \mathcal{D}\cup\mathcal{O}}^{} p(i|X) * p(j|i,X) 
          = \sum_{ i \in \mathcal{D}\cup\mathcal{O}}^{} \hat{A}_{anchor}(i) * \hat{A}_{\mathcal{G}_i}(j),
\label{eq:hie_regular_pred}
\end{equation}
where $j \in \mathcal{R}$.
Thus, 
instead of \Eref{eq:modified},
we obtain the final action probability predictions for our hierarchical architecture $\hat{A}(a) = p(a|X)$ as 
\begin{equation}
    \hat{A}(a) = 
\begin{cases}
    \hat{A}_{anchor}(a),
    & \text{if } a\in \mathcal{D}\\
    \sum_{ i \in \mathcal{D}\cup\mathcal{O}}^{} \hat{A}_{anchor}(i) * \hat{A}_{\mathcal{G}_i}(a),
    & \text{otherwise.}
\end{cases}
\label{eq:hie_action_pred}
\end{equation}
We use the same method as in \Eref{eq:base_hoi_pred} and cross-entropy loss to compute the final HOI probability prediction $\hat{Y}$ and the corresponding loss $\mathcal{L}$.

To demonstrate the effectiveness of the hierarchical learning, we introduce another two baselines,  \emph{\textbf{MultiTask}} and \emph{\textbf{TwoStream}}, that lie between the \emph{\textbf{Modified Baseline}} and our hierarchical learning. \emph{\textbf{MultiTask}} only uses the anchor action classification as an additional multi-task element to the \emph{\textbf{Modified Baseline}}. %
\emph{\textbf{TwoStream}} separately predicts the anchor and the regular actions
but without using the hierarchical modeling between anchor and regular actions. 
\subsection{{ACP Projection for Knowledge Distillation}}
\label{sec.knowledge_distillation}



Knowledge distillation~\cite{hinton2015distilling} was originally proposed to transfer knowledge from a large network to a smaller one. 
{Recently, knowledge distillation has been utilized for various purpose such as life-long learning~\cite{li2016learning} or multi-task learning~\cite{kim2018disjoint}.}
Hu~\etal~\cite{hu2016harnessing} extended this concept to distill prior knowledge in the form of logic rules into a deep neural network. Specifically, they propose a teacher-student framework to project the network prediction (student) to a rule-regularized subspace (teacher), where the process is termed distillation. 
The network is then updated to balance between emulating the teacher's output and predicting the true labels.

Our work fits this setting because the ACPs can act as a prior to distill. 
We first introduce an \emph{ACP Projection} to map the action distribution to the ACP constraints. Then, we use the teacher-student framework~\cite{hu2016harnessing} to distill knowledge from ACPs.


\noindent\textbf{ACP Projection} In ACP Projection, an arbitrary action distribution 
$A = \{p(i), i \in [0, N)\} \in \mathbb{R}^{ N}$
is projected into the ACP-constrained probability space:
\begin{equation}
A^{*} = project(A, C, C^{'}) \in \mathbb{R}^{N},
\end{equation}
where $A^{*}$ is the projected action prediction. The projected probability for the $j$-th action $A^{*}(j) = p(j^{*})$ is generated using the law of total probability:
\begin{equation}
    p(j^{*})  =\dfrac{1}{N}\sum_{i=1}^{N} (p(i) * p(j|i) + p(i^{'}) * p(j|i^{'})) 
          = \dfrac{1}{N} (\sum_{i=1}^{N} p(i) * c_{ij} + \sum_{i=1}^{N} (1-p(i)) * c_{ij}^{'}). 
\end{equation}


In matrix form, the ACP projection is expressed as
\begin{equation}
project(A, C, C^{'}) = \dfrac{AC + (1-A)C^{'}}{N}.
\end{equation}

In practice, we use the object-based action co-occurrence matrices $C_{o}\in\mathbb{R}^{N\times N}$ and $C_{o}^{'}\in\mathbb{R}^{N\times N}$ which only count actions related to a specific object $o$. \Fref{fig:co-occurrence} shows examples of $C_{o}$ with respect to object classes.
Also, we give different weights $\alpha$ and $\beta$ {as hyper-parameters} to the action co-occurrence matrix $C_{o}$ and its complementary matrix $C_{o}^{'}$, with the weights subject to $\alpha + \beta = 2,\alpha > \beta$.
The projection function is then modified as
\begin{equation}
project(A, C_{o}, C^{'}_{o}) = \dfrac{\alpha AC_{o} + \beta (1-A)C^{'}_{o}}{N}.
\end{equation}
This is done because we empirically found the co-occurrence relationships in $C_{o}$ to generally be much stronger then the complementary actions in $C^{'}_{o}$.

\noindent\textbf{Teacher-Student Framework} Now we can distill knowledge from the ACPs using ACP Projection in both the training and inference phases. There are three ways ACP Projection can be used:
(1) Directly project the action prediction $\hat{A}$ into the ACP-constrained probability space at the testing phase to obtain the final action output (denoted as \emph{\textbf{PostProcess}}). 
(2) Project the action prediction $\hat{A}$ in the training phase and use the projected action as an additional learning target~\cite{hu2016harnessing,yu2017visual}.
(3) Project the ground truth label $H^{gt}$, $O^{gt}$, and $A^{gt}$\footnote{The triplet ground truth labels $H^{gt}$, $O^{gt}$, and $A^{gt}$ are straightforward to determine from the HOI ground truth label $Y^{gt}$.} to the ACP space in the training phase and use the projected action $project(A^{gt}, C_{{O}^{gt}}, C_{{O}^{gt}}^{'})$ as an additional learning target.
The second and third items are incorporated into the teacher-student framework as terms in a new objective function {(denoted as \emph{\textbf{Distillation}})}:
\begin{equation}
\mathcal{L}_{total} = \lambda_1 \mathcal{L}(\hat{Y}, Y^{gt}) + \lambda_2 \mathcal{L}(\hat{Y}, \hat{Y}_{projO}) + \lambda_3 \mathcal{L}(\hat{Y}, Y^{gt}_{projO}),
\label{eqn:total_loss}
\end{equation}
where
\begin{equation}
\hat{Y}_{projO} = joint(\hat{H}, \hat{O}, project(\hat{A}, C_{\hat{O}}, C_{\hat{O}}^{'}))  \in \mathbb{R}^{M},
\label{eqn:teacher2_self}
\end{equation}
\begin{equation}
Y^{gt}_{projO} = joint({H}^{gt},{O}^{gt},project(A^{gt}, C_{{O}^{gt}}, C_{{O}^{gt}}^{'}))  \in \mathbb{R}^{M}.
\label{eqn:teacher2_gt}
\end{equation}
$\lambda_1, \lambda_2, \lambda_3$ are balancing weights among the ground truth HOI term and the teacher objectives.
The object type can be easily determined from the object probability predictions $\hat{O}$ or the ground truth label ${O}^{gt}$.
\section{Experiments}
\vspace{-2mm}

{The goal of the experiments is to show the effectiveness and generalizability of our method.
In particular, we show that our method can consistently alleviate the long-tailed distribution problem in various setups by improving performance especially for rare HOI classes.} 
In this section, we describe the experimental setups, competing methods and provide performance evaluations of HOI detection.

\subsection{Datasets and Metrics}
\label{sec.dataset}
We evaluate the performance of our model on the two popular HOI detection benchmark datasets, \textbf{HICO-Det}~\cite{chao2018learning} and \textbf{V-COCO}~\cite{gupta2015visual}.
HICO-Det~\cite{chao2018learning} extends the HICO (Humans Interacting with Common Objects) dataset~\cite{chao2015hico} which contains 600 human-object interaction categories for 80 objects. 
Different from the HICO dataset, HICO-Det additionally contains bounding box annotations for humans and objects of each HOI category. 
The vocabulary of objects matches the 80 categories of MS COCO~\cite{lin2014microsoft}, and there are 117 different verb (action) categories. 
The number of all possible triplets is $117\times80$, but the dataset contains positive examples for only 600 triplets. 
The training set of HICO-Det contains 38,118 images and 117,871 HOI annotations for 600 HOI classes. 
The test set has 9,658 images and 33,405 HOI instances.

For evaluation, HICO-Det uses the mean average precision (mAP) metric.
Here, an HOI detection is counted as a true positive if the minimum of the human overlap IOU and object overlap
IOU with the ground truth is greater than 0.5. 
Following \cite{chao2018learning}, HOI detection performance is reported for three different HOI category sets: (1) all 600 HOI categories (Full), (2) 138 categories with fewer than 10 training samples (Rare), and (3) the remaining 462 categories with more than 10 training samples (Non-rare).

V-COCO (Verbs in COCO) is a subset of MS-COCO~\cite{lin2014microsoft}, which consists of 10,346 images (2,533, 2,867, 4,946 for training, validation and test, respectively) and 16,199 human instances. 
Each person is annotated with binary labels for 26 action classes.
For the evaluation metric, same as for evaluation on HICO-Det, we use the AP score.

\begin{table}[t]
\centering
    \begin{minipage}{0.53\textwidth}
    \centering
        \caption{Ablation study on the HICO-Det dataset. Our final model that includes both hierarchical architecture and distillation followed by post processing (Ours, ACP) shows the best performance among the baselines.}
        \vspace{-2mm}
        \resizebox{1\linewidth}{!}{
            \begin{tabular}{l| ccc}
        		\hline
        			&		Full&Rare&Non-rare\\
        		\hline
        		\hline
        		Baseline	  & 17.56&   	13.23&   18.85 \\
        		Modified Baseline &   19.09	& 13.09  & 20.89\\
        		\hline
        		+Hierarchical only &   20.03	&14.52   &21.67\\
        		+Distillation only &   19.98	&13.67   &21.86\\
        		+Hierarchical+Distillation &   {20.25}	& {15.33}   &{21.72}\\
        		\hline
        		+Hierarchical+Distillation+Post (Ours, ACP)&   \textbf{20.59}	& \textbf{15.92}   & \textbf{21.98}\\
        		\hline
    		\end{tabular}
		}
	\label{table:ablation}
    \end{minipage}%
    \hfill
    \begin{minipage}{0.43\textwidth}
    \centering
        \caption{Performance of our models with different architectures for action prediction module. Our 
         model ((D) Hierarchical) shows the best performance among the design choices.
    	\vspace{-2mm}
    	}
        \resizebox{1\linewidth}{!}{
            \begin{tabular}{l| ccc}\hline
			&		Full&Rare&Non-rare\\\hline\hline
			(A) Modified Baseline &   19.09	& 13.09  & 20.89\\
			(B) MultiTask &   19.54	&13.93   &21.22\\
			(C) TwoStream &   19.63	& 13.67   & 21.41\\
			(D) Hierarchical &   \textbf{20.03}	&\textbf{14.52}   &\textbf{21.67}\\
			\hline
		\end{tabular}
    	}
	\label{table:architecture}	
    \end{minipage}
    \vspace{-6mm}
\end{table}


\subsection{Quantitative Results}
\noindent\textbf{Ablation study}
In the ablations, the `No-Frills' baseline network~\cite{gupta2019no} we used is denoted as the \emph{\textbf{Baseline}}. We first evaluate the effectiveness of the core design components in the proposed method including (1) our simple modification to \emph{\textbf{Baseline}} in \Sref{sec.architecture}, denoted as \textbf{\emph{Modified Baseline}}; (2) the hierarchical learning technique introduced in \Sref{sec.architecture}, denoted as \emph{\textbf{Hierarchical}}; and (3) the knowledge distillation technique presented in {\Eref{eqn:total_loss}} of \Sref{sec.knowledge_distillation}, denoted as \emph{\textbf{Distillation}}. \Tref{table:ablation} gives a comprehensive evaluation for each component. We draw conclusions from it one-by-one.

\emph{\textbf{First, our baseline network is strong.}} Our \emph{\textbf{Modified Baseline}} achieves 19.09 mAP and surpasses the `No-Frills' \emph{\textbf{Baseline}} by 1.51 mAP (a relative 8.7\% improvement), which is already competitive to the state-of-the-art result~\cite{peyre2019detecting} and serves as a strong baseline. 

\emph{\textbf{Second, both hierarchical learning and knowledge distillation are effective.}} This is concluded by adding \emph{\textbf{Hierarchical}} and \emph{\textbf{Distillation}} to the \emph{\textbf{Modified Baseline}}, respectively. Specifically, \emph{\textbf{+Hierarchical}} improves the modified baseline by 0.94 mAP (a relative 4.9\% improvement), and \emph{\textbf{+Distillation}} {({training with \Eref{eqn:total_loss}})} improves the modified baseline by 0.89 mAP (a relative 4.7\% improvement). Including both obtains 1.16 mAP improvement (relatively better by 6.1\%).

\emph{\textbf{Third, the proposed ACP method achieves a new state-of-the-art.}} Our final result is generated by further using the \emph{\textbf{PostProcess}} step (introduced in \Sref{sec.knowledge_distillation}) that projects the final action prediction into the ACP constrained space. Our method achieves 20.59 mAP (relative 7.9\% improvement) for Full HOI categories, 15.92 mAP (relative 21.6\% improvement) for Rare HOI categories, and 21.98 mAP (relative 5.2\% improvement) for Non-rare HOI categories. 
Note that our method made especially significant improvements for Rare classes, which supports the claim that the proposed method can alleviate the long-tailed distribution problem of HOI detection datasets.
This result sets the new state-of-the-art on both the HICO-Det and V-COCO datasets as shown in \Tref{table:hico} and \Tref{table:vcoco}.



In addition, the \textbf{\emph{MultiTask}},\textbf{\emph{TwoStream}}, and our \textbf{\emph{Hierarchical}} architecture are compared in \Tref{table:architecture}. From \textbf{\emph{MultiTask}}, it can be seen that the \emph{softmax} based anchor action classification already brings benefits to the \emph{\textbf{Modified Baseline}} when used only in a multi-task learning manner. 
{From \textbf{\emph{TwoStream}}, separately modeling the anchor and the regular classes leads to a slight more improvements compared to \textbf{\emph{MultiTask}}}.
Moreover, our \textbf{\emph{Hierarchical}} architecture  improves upon \textbf{\emph{TwoStream}} by explicitly modeling the hierarchy between anchor and regular action predictions.


\noindent\textbf{Comparison with the state-of-the-art}
We compare our method with the previous state-of-the-art techniques in \Tref{table:hico}. 
Among the methods included in this comparison are the benchmark model of the HICO-Det dataset~\cite{chao2018learning}, the baseline model that we modified from~\cite{gupta2019no}, and the current published state-of-the-art method~\cite{peyre2019detecting}. 
As shown in \Tref{table:hico}, our final model (Ours) shows significant improvements over our baseline model on all metrics, and shows favorable performance against the current state-of-the-art model in terms of all the metrics. 
In particular, our model surpasses the current state-of-the-art model~\cite{peyre2019detecting} by 1.19 mAP.


\begin{table}[t]
\centering
    \begin{minipage}{0.53\textwidth}
    \centering
        \caption{Results on the HICO-Det dataset compared to the previous state-of-the-art methods. Our model shows favorable performance against the current state-of-the-art models on all the metrics.}
        \vspace{0mm}
        \resizebox{1\linewidth}{!}{
            \begin{tabular}{l| ccc}\hline
    			&		Full&Rare&Non-rare\\\hline\hline
    			Shen~\etal~\cite{shen2018scaling}	            & 6.46	& 4.24  & 7.12\\
    			HO-RCNN~\cite{chao2018learning}                 & 7.81	& 5.37  & 8.54\\
    			Gupta~\etal~\cite{gupta2015visual} impl. by~\cite{gkioxari2018detecting}  &   9.09	&7.02   &9.71\\
    			InteractNet~\cite{gkioxari2018detecting}	    & 9.94	& 7.16  & 10.77\\
    			GPNN~\cite{qi2018learning}	                    & 13.11 & 9.34  & 14.23\\
    			iCAN~\cite{gao2018ican}	                        & 14.84 & 10.45 & 16.15\\
    			iHOI~\cite{xu2019interact}                      & 13.39 & 9.51  & 14.55 \\
    			with Knowledge~\cite{xu2019learning}	        & 14.70 & 13.26 & 15.13\\
    			Interactiveness Prior~\cite{li2019transferable} & 17.22 & 13.51 & 18.32\\
    			Contextual Attention~\cite{wang2019deep}        & 16.24 & 11.16 & 17.75 \\
    			No-Frills~\cite{gupta2019no}	                & 17.18 & 12.17 & 18.68\\
    			RPNN~\cite{zhou2019relation}                    & 17.35 & 12.78 & 18.71\\
    			PMFNet~\cite{wan2019pose}                       & 17.46 & 15.65 & 18.00\\
    			Peyre~\etal~\cite{peyre2019detecting}	        & 19.40 & 15.40 & 20.75\\
    			\hline
    			{Our baseline}	  &   17.56	&13.23   &18.85\\
    			\textbf{ACP (Ours)}	  &   \textbf{20.59}	& \textbf{15.92}   & \textbf{21.98}\\
    			\hline
    		\end{tabular}
		}
		\label{table:hico}
    \end{minipage}%
    \hfill
    \begin{minipage}{0.43\textwidth}
    \centering
        \caption{Results on the V-COCO dataset. {For our method, we show results both for constructing the ACP from V-COCO and for using the ACP constructed from HICO-Det instead. Both of these models show favorable performance against the current state-of-the-art models.}}
        \vspace{-1mm}
        \resizebox{.85\linewidth}{!}{
            \begin{tabular}{l| c}\hline
        			&		$AP_{role}$\\
        			\hline\hline
        			Gupta~\etal~\cite{gupta2015visual} impl. by~\cite{gkioxari2018detecting}	  &   31.8	\\
        			InteractNet~\cite{gkioxari2018detecting}	  &   40.0	\\
        			GPNN~\cite{qi2018learning}	  &   44.0\\
        			iCAN~\cite{gao2018ican}	  &   45.3\\
        			iHOI~\cite{xu2019interact} & 45.79 \\
        			with Knowledge~\cite{xu2019learning}	  &   45.9\\
        			Interactiveness Prior~\cite{li2019transferable}	  &   48.7\\
        			Contextual Attention~\cite{wang2019deep} & 47.3\\
        			RPNN~\cite{zhou2019relation}&47.53\\
        			PMFNet~\cite{wan2019pose} & 52.0 \\
        			\hline
        			Our baseline            &       48.91   \\
        			\textbf{ACP (Ours, V-COCO)}	  &   \textbf{52.98}	\\
        			\textbf{ACP (Ours, HICO-Det)}	  &   \textbf{53.23}	\\
        			\hline
        	\end{tabular}
        	}
        \label{table:vcoco}
    \end{minipage}
    \vspace{-6mm}
\end{table}

\noindent\textbf{Results on V-COCO dataset}
To show the generalizability of our method, we also evaluate our method on the V-COCO dataset.
Note that the exact same method is directly applied to both HICO-Det and V-COCO, including the co-occurrence matrix, anchor action selection, and the architecture design.
{We also constructed a co-occurrence matrix from V-COCO, but the matrix was sparse. Thus, to better take advantage of our idea, we instead use the co-occurrence matrix collected from the HICO-Det dataset to train on the V-COCO.}
\Tref{table:vcoco} shows the performance of our model (Ours, HICO-Det) compared to the recent state-of-the-art HOI detectors on the V-COCO dataset.
{In addition, we show results of our model with the co-occurrence matrix constructed from the V-COCO dataset (Ours, V-COCO).}
{Both of these models show} favorable performance on the V-COCO dataset against the previous state-of-the-art model~\cite{wan2019pose}.

\begin{table}[t]
\centering
\vspace{0mm}
    \caption{Results on the zero-shot triplets of the HICO-Det dataset. 
    Our final model shows better performance than Peyre~\etal by a large margin.
    Note that our ACP model under the zero-shot setting even outperforms the supervised setting of our baseline.
    }
    \vspace{-2mm}
    \resizebox{.5\linewidth}{!}{%
		\begin{tabular}{l| c}\hline
			&		mAP\\\hline\hline
			Peyre~\etal~\cite{peyre2019detecting} Supervised & 33.7\\
			Peyre~\etal~\cite{peyre2019detecting} Zero-Shot &   24.1\\
			Peyre~\etal~\cite{peyre2019detecting} Zero-Shot with Aggregation &   28.6\\
			\hline
			Ours Supervised (Modified Baseline) & 33.27\\
			Ours Zero-Shot (Modified Baseline) &   20.34	\\
			\textbf{Ours Zero-Shot (ACP)}
			&   \textbf{34.95}\\
			\hline
		\end{tabular}
    }
    \vspace{-4mm}
	\vspace{0mm}
	\label{table:zero-shot}	
\end{table}

\noindent\textbf{Results of the zero-shot setup on the HICO-Det dataset}
The \emph{zero-shot setting} on the HICO-Det dataset is defined by Peyre~\etal~\cite{peyre2019detecting}. 
Specifically, we select a set of 25 HOI classes that we treat as unseen classes and exclude them and their labels in the training phase. 
However, we still let the model predict those 25 unseen classes in the test phase, which is known as the zero-shot problem. 
These HOI classes are randomly selected among the set of non-rare HOI classes.
Since Peyre~\etal did not provide which specific HOI classes they selected, we select the unseen HOI classes such that the performance (mAP) for these classes in our \textbf{\emph{Modified Baseline}} model (introduced in \Sref{sec.architecture}) is similar to the corresponding \emph{\textbf{Supervised}} baseline in~\cite{peyre2019detecting}. 
In \Tref{table:zero-shot}, we show results of our final model (ACP) and our modified baseline model compared to the corresponding setting reported in~\cite{peyre2019detecting}.
Our final model shows better performance (35.0 mAP) than Peyre~\etal (28.6 mAP) by a large margin (relative 22.4\% improvement).
This result is remarkable in that our ACP model under the zero-shot setting even \emph{outperforms} the supervised setting of our baseline model, indicating the power of the proposed ACP method to effectively leverage prior knowledge on action co-occurrences. Furthermore, the analogy transfer method proposed by Peyre~\etal (denoted as aggregation) requires large-scale linguistic knowledge to train a word representation, whereas our model only requires the co-occurrence information of the labels in the dataset, which is much easier to obtain.
We conclude that the proposed method is effective for the zero-shot problem while being easy to implement.

\begin{figure}[t]
\centering
\vspace{0mm}
    \begin{minipage}{0.48\textwidth}
    \centering
    \includegraphics[width=.8\linewidth,keepaspectratio]{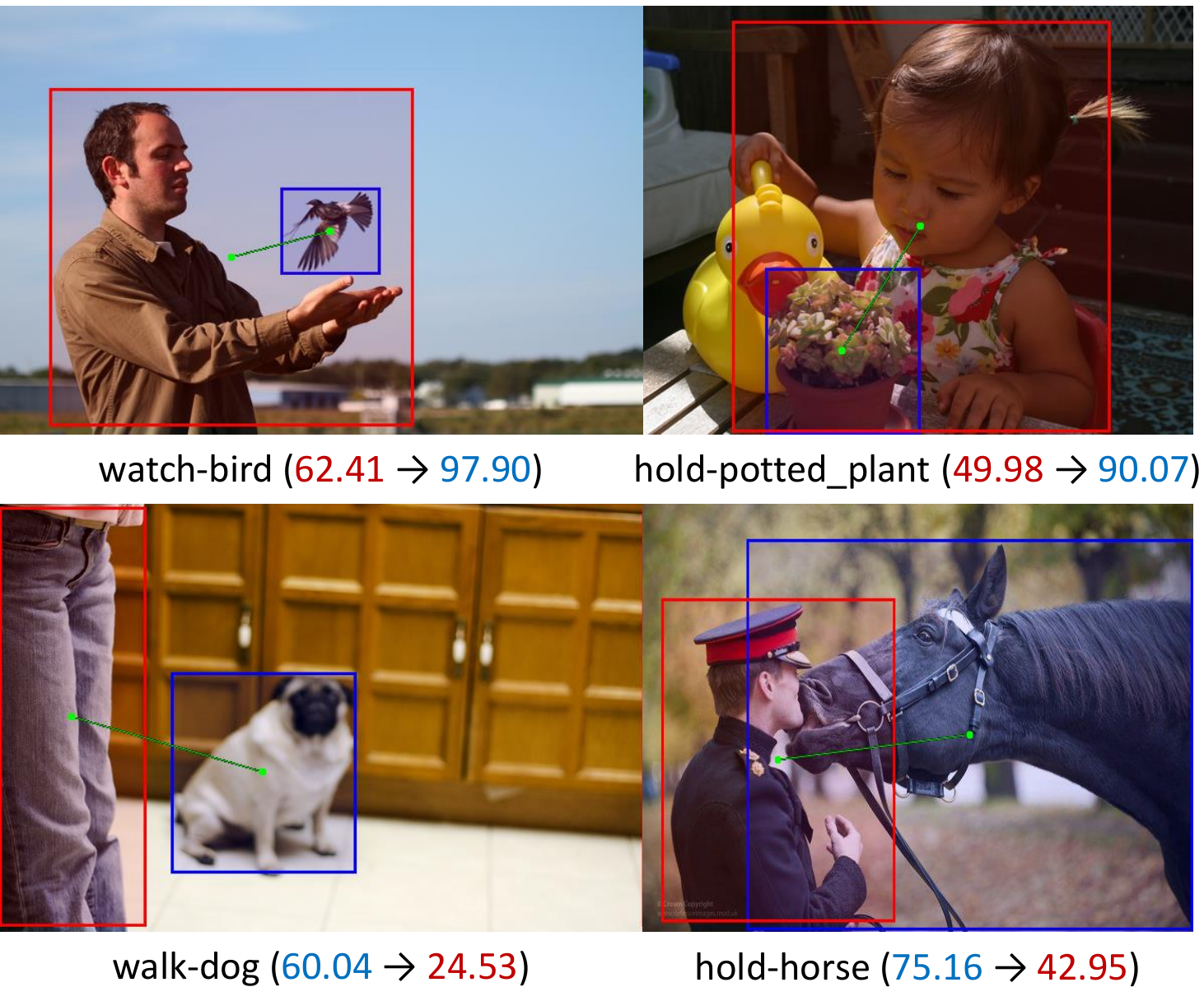}
	\vspace{-2mm}
	\caption{The HOI probability before and after applying the projection function $project(\cdot)$ on our model's HOI prediction (\emph{\textbf{PostProcess}}). 
	Note that \emph{\textbf{PostProcess}} can be done without any optimization.
	}
	\label{fig:prob_change}
    \end{minipage}%
    \hfill
    \begin{minipage}{0.48\textwidth}
    \centering
    \includegraphics[width=.7\linewidth,keepaspectratio]{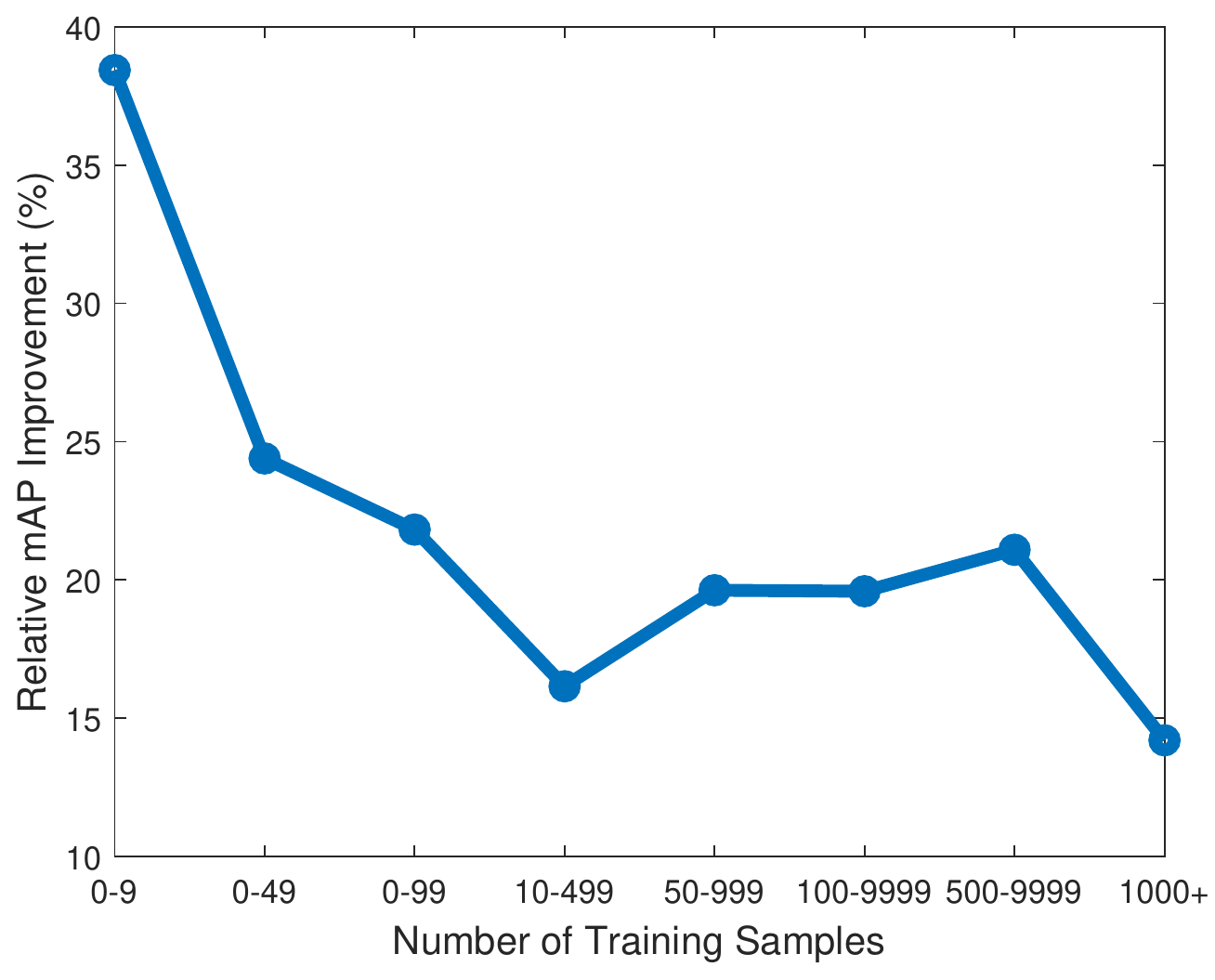}\\
    \vspace{-2mm}
	\caption{The relative mAP score improvements for various HOI sets with the different number of training samples. Our method is able to improve the performance especially when the number of training samples is small (38.24\% improvement for 0-9 samples).
	}
	\label{fig:num_of_sample}
    \end{minipage}
    \vspace{-4mm}
\end{figure}

\subsection{Additional Analysis}
\noindent\textbf{Score Improvement after ACP Projection}
We also show the HOI probability change from before to after applying the projection function $project(\cdot)$ on our model's HOI prediction (\ie~the effect of \emph{\textbf{PostProcess}} introduced in \Sref{sec.knowledge_distillation}) in \Fref{fig:prob_change}.
Leveraging co-occurrence matrix $C$ can not only increase the score for true classes (top) but also reduce the score for false classes (bottom). 
Note that this change can be achieved \emph{without any optimization} process.

\noindent\textbf{Performance on various sets with different number of training samples}
{Finally, in \Fref{fig:num_of_sample}, we show the relative mAP score improvements of our model compared to the baseline model by computing mAP on various sets of HOI classes that have different number of training samples.
Our method shows positive performance improvements for all numbers of training samples. 
Also, there is a trend that HOI classes with a small number of training samples mostly show larger performance improvements.
In particular, for HOI classes with the number of training samples between 0 and 9, our model achieves 38.24\% improvement compared to the baseline model.
These results indicate that the proposed method is able to improve the performance of an HOI detector, especially for classes with few training samples.
}


\vspace{-2mm}

\section{Conclusion}
We introduced a novel method to effectively train an HOI detector by leveraging prior knowledge on action co-occurrences in two different ways, via the architecture and via the loss function.
Our proposed method consistently achieves favorable performance compared to the current state-of-the-art methods in various setups.
Co-occurrence information not only is helpful for alleviating the long-tailed distribution problem but also can be easily obtained. 
{A direction for future work is to construct and utilize co-occurrence priors for other relationship-based vision tasks~\cite{johnson2017clevr,kim2019dense,lu2016visual}. }

\vfill

\noindent{\textbf{Acknowledgements.}}
This work was supported by the Institute for Information \& Communications Technology Promotion (2017-0-01772) grant funded by the Korea government.

\clearpage
%
%
\bibliographystyle{splncs04}
\bibliography{egbib}

\clearpage

\end{document}